\documentclass[conference]{IEEEtran}
\IEEEoverridecommandlockouts
\usepackage{cite}
\usepackage{amsmath,amssymb,amsfonts}
\usepackage{algorithmic}
\usepackage{graphicx}
\usepackage{textcomp}
\usepackage{xcolor}
\def\BibTeX{{\rm B\kern-.05em{\sc i\kern-.025em b}\kern-.08em
    T\kern-.1667em\lower.7ex\hbox{E}\kern-.125emX}}

\usepackage{booktabs}	
\usepackage{multirow,multicol}
\usepackage{subfigure}
\usepackage[flushleft]{threeparttable}
\usepackage{lipsum}
\usepackage{siunitx}
\usepackage[utf8]{inputenc}
\usepackage[cjk]{kotex}
\usepackage[textsize=tiny]{todonotes}

\begin{document}

\title{Long-term Time Series Forecasting based on Decomposition and Neural Ordinary Differential Equations}

\author{
    \IEEEauthorblockN{
        Seonkyu Lim\IEEEauthorrefmark{1}\IEEEauthorrefmark{2}\textsuperscript{\textsection},
        Jaehyeon Park\IEEEauthorrefmark{1}\textsuperscript{\textsection}, 
        Seojin Kim\IEEEauthorrefmark{1}\textsuperscript{\textsection}, 
        Hyowon Wi\IEEEauthorrefmark{1}, 
        Haksoo Lim\IEEEauthorrefmark{1}, \\
        Jinsung Jeon\IEEEauthorrefmark{1}, 
        Jeongwhan Choi\IEEEauthorrefmark{1} and
        Noseong Park\IEEEauthorrefmark{1}
    }
    \IEEEauthorblockA{
        \textit{Yonsei University}\IEEEauthorrefmark{1}\textit{, Seoul, South Korea}
    }
    \IEEEauthorblockA{
        \textit{Korea Financial Telecommunications \& Clearings Institute}\IEEEauthorrefmark{2}\textit{, Seoul, South Korea}
    }
    \IEEEauthorblockA{
        sklim@kftc.or.kr\IEEEauthorrefmark{2}, 
        \{jaehyun9907, bwnebs1, wihyowon, limhaksoo96, jjsjjs0902, jeongwhan.choi, noseong\}@yonsei.ac.kr\IEEEauthorrefmark{1}
    }
}

% For papers in which all authors are employed by the US government, the copyright notice is: U.S. Government work not protected by U.S. copyright
% For papers in which all authors are employed by a Crown government (UK, Canada, and Australia), the copyright notice is: 979-8-3503-2445-7/23/$31.00 ©2023 Crown
% For papers in which all authors are employed by the European Union, the copyright notice is: 979-8-3503-2445-7/23/$31.00 ©2023 European Union
% For all other papers the copyright notice is: 979-8-3503-2445-7/23/$31.00 ©2023 IEEE

\IEEEoverridecommandlockouts
\IEEEpubid{\makebox[\columnwidth]{979-8-3503-2445-7/23/\$31.00~\copyright2023 IEEE \hfill}
\hspace{\columnsep}\makebox[\columnwidth]{ }}

\maketitle

\begingroup\renewcommand\thefootnote{\textsection}
\footnotetext{These authors contributed equally to this research.}
\endgroup

\IEEEpubidadjcol

\begin{abstract}
Long-term time series forecasting (LTSF) is a challenging task that has been investigated in various domains such as finance investment, health care, traffic, and weather forecasting. 
In recent years, 
Linear-based LTSF models showed better performance, pointing out the problem of Transformer-based approaches causing temporal information loss. However, Linear-based approach has also limitations that the model is too simple to comprehensively exploit the characteristics of the dataset.
To solve these limitations, we propose LTSF-DNODE, which applies a model based on linear ordinary differential equations (ODEs) and a time series decomposition method according to data statistical characteristics.
We show that LTSF-DNODE outperforms the baselines on various real-world datasets. In addition, for each dataset, we explore the impacts of regularization in the neural ordinary differential equation (NODE) framework.
\end{abstract}

\begin{IEEEkeywords}
long-term time series forecasting, time series decomposition, instance normalization, neural ordinary differential equations
\end{IEEEkeywords}

\section{Introduction}
Time series data is continuously generated from various real-world applications. To utilize this data, numerous approaches have been developed in the fields such as forecasting~\cite{lim2021time,wen2022transformers,hwang2021climate,choi2022graph,jhin2021attentive,jhin2022exit,choi2023graph,choi2023climate,hong2022timekit}, classification~\cite{alqahtani2021deep,wen2022transformers,ismail2020inceptiontime,chen2021multi,jhin2023learnable,jhin2021attentive,jhin2022exit}, and generation~\cite{jeon2022gt,yoon2019time,alaa2021generative,donahue2018adversarial}. Among them, time series forecasting is one of the most important research topics in deep learning. For time series forecasting, recurrent neural network (RNN)-based models were used, such as LSTM~\cite{hochreiter1997long} and GRU~\cite{cho2014properties}. These models performed well in time series forecasting but suffered from error accumulation due to their iterative multi-step forecasting approach, especially when dealing with long-term time series forecasting (LTSF).

To overcome this challenge, there have been various attempts in the past few years. Among them, the Transformer-based approaches~\cite{wen2022transformers}, which enable direct multi-step forecasting, show significant performance improvement. Transformer~\cite{vaswani2017attention} has demonstrated remarkable performance in various natural language processing tasks, and its ability to effectively capture long-range dependencies and interactions in sequential data makes it suitable for application in LTSF. 

Despite the impressive achievements of Transformer-based models in LTSF, they have struggled with temporal information loss caused by the self-attention mechanism as a result of permutation invariant and anti-order properties~\cite{zeng2023transformers}.
Furthermore, it was demonstrated that the error upper bound of Transformer-based models, which is one of the non-linear deep learning approaches, is higher than linear regression~\cite{li2022simpler}.

Recently, simple Linear-based methods~\cite{zeng2023transformers, li2022simpler}, as non-Transformer-based models, show better performance compared to complex Transformer-based models.
These approaches are novel attempts in LTSF where Transformer architecture has recently taken the lead. However, they have limits to comprehensively exploit the intricate characteristics of the time series dataset as the models are too simple.

Inspired by these insights, we aim to design a model that takes into account the characteristics of each dataset by introducing some sophistication to the model architecture. Our proposed model, LTSF-DNODE, adeptly harnesses temporal information by employing time series decomposition and the neural ordinary differential equation (NODE) framework. 
The NODE framework transforms a single linear layer into time-derivative modeling. It uses a structure composed of the same single linear layer, but can better capture the complex dynamics of time series data, providing various advantages within time series processing.

Our contributions can be summarized as follows:

\begin{itemize}
    \item We analyze the temporal information of each real-world time series datasets with exploratory data analysis. This investigation helps us detect the presence of seasonality, guiding us to perform decomposition appropriately.
    \item Based on the NODE framework, we demonstrate the following benefits of time-derivative modeling:
    i) It is suitable for time series tasks by interpreting discrete linear layers as continuous linear layers, and ii) more advanced regularization is available.
    \item Through various empirical experiments, we demonstrate that the NODE framework and decomposition according to data characteristics are effective in LTSF.
\end{itemize}

\section{Preliminaries}
In this section, we review the decomposition method and neural ordinary differential equations. Following that, we conduct empirical explorations to investigate the effectiveness of these techniques on LTSF.

\subsection{Problem Formulation}
The objective of LTSF is to forecast from an input sequence of historical time series data to a corresponding future sequence. Given the input historical data $\mathbf{X} = [\mathbf{x}_1, \ldots, \mathbf{x}_L ]^{\textsf{T}}\in\mathbb{R}^{L\times F}$, LTSF models forecast $\mathbf{Y} = [\mathbf{y}_1, \ldots, \mathbf{y}_H ]^{\textsf{T}}\in\mathbb{R}^{H\times F}$, where $L$ is the look-back window size, $H$ is the forecasting horizon and $F$ is the feature dimension. The LTSF problem deals with cases where $H$ is longer than 1, and  $F$ is not restricted to univariate cases.

\subsection{Time Series Decomposition}
Data preprocessing is used to enhance data quality as an input to the LTSF method, enabling it to deliver better outcomes. According to \cite{salles2019nonstationary}, suitable preprocessing methods for non-stationary time series increased forecasting accuracy by more than 10\% on over 95\% of the temporal data.

Time series decomposition is a pivotal technique in time series preprocessing~\cite{hyndman2018forecasting}. The decomposition methods (e.g., STL~\cite{cleveland1990stl} and SEAT~\cite{dagum2016seasonal}) typically decompose the time series into three components: trend, seasonality, and residual component. The decomposition can be represented as follows:
\begin{align}
    \mathbf{X} = \mathbf{T} + \mathbf{S} + \mathbf{R},
\end{align}
where $\mathbf{T},\mathbf{S},\mathbf{R}\in\mathbb{R}^{L\times F}$ respectively represent the trend, seasonality, and residual component.
The trend is a general systematic linear or nonlinear component that changes over time and does not repeat within the given timeframe, usually identified by the two-sided simple moving average~\cite{hyndman2011moving}. Seasonality means a pattern with a particular cycle in a time series. To extract seasonality, the classic additive decomposition method averages the detrended series $(\mathbf{X}-\mathbf{T})$ across a predetermined period. Residual is the remaining value after removing the trend and seasonality from a series.

We use this method to provide an overall understanding of time series datasets with exploratory data analysis.

\subsection{Neural Ordinary Differential Equations}

Neural ordinary differential equations (NODEs)~\cite{chen2018neural} can handle time series data in a continuous manner, using the differential equation as follows:
\begin{align}
    \mathbf{z}(T) &= \mathbf{z}(0) + \int_{0}^{T} f(\mathbf{z}(t),t;\boldsymbol{\theta}_f) dt, \label{eq:ode}
\end{align} where $f(\mathbf{z}(t),t;\boldsymbol{\theta}_f)$, called an ODE function, is a neural network to approximate the derivative of $\mathbf{z}(t)$ with respect to $t$ (denoted as $\frac{d\mathbf{z}(t)}{dt}$). 

\begin{figure}[t]
    \centering
    \subfigure[Euler method]{\includegraphics[width=0.3\columnwidth]{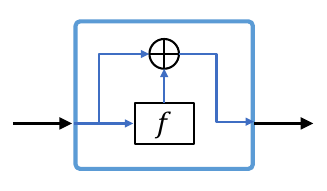}}
    \subfigure[RK4 method]{\includegraphics[width=0.53\columnwidth]{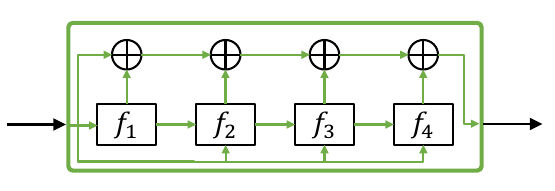}}
    \caption{The (a) explicit Euler method and (b) RK4 method in a step. We note that the Euler method creates the residual connection and RK4 makes the dense connection given the ODE function $f$ parameterized by $\theta_f$.}
    \label{fig:solver}
\end{figure}
To solve the integral problem, the NODEs use the ODE solver. The ODE solvers divide the integral time domain $[0, T]$ in Eq.~\eqref{eq:ode} into small steps and convert the integral into many steps of additions. For example, one step of the explicit Euler method, a typical ODE solver, is as follows:
\begin{align}
    \mathbf{z}(t + s) = \mathbf{z}(t) + s \cdot f(\mathbf{z}(t), t;\boldsymbol{\theta}_f),
\end{align} where $s \in (0,1]$ is step size of the Euler method. A more sophisticated method, such as the 4-th order Runge-Kutta (RK4) method, is as follows:
\begin{align}\label{eq:rk4}
\mathbf{z}(t + s) = \mathbf{z}(t) + \frac{s}{6}\Big(f_1 + 2f_2 + 2f_3 + f_4\Big),
\end{align}where $f_1 = f(\mathbf{z}(t), t;\boldsymbol{\theta}_f)$, $f_2 = f(\mathbf{z}(t) + \frac{s}{2}f_1, t+\frac{s}{2};\boldsymbol{\theta}_f)$, $f_3 = f(\mathbf{z}(t) + \frac{s}{2}f_2, t+\frac{s}{2};\boldsymbol{\theta}_f)$, and $f_4 = f(\mathbf{z}(t)+sf_3, t+s;\boldsymbol{\theta}_f)$. 

Fig.~\ref{fig:solver} denotes the Euler and RK4 methods. These are some of the explicit methods that have a fixed step size. The RK4 method requires four times as much work as the Euler method in a single step. When the step size is 1, the Euler method is equivalent to the residual connection. Similarly, when $f$ represents a neural network layer, the RK4 method is analogous to the dense connection. 
On the other hand, one of the most advanced methods, Dormand-Prince (DOPRI)~\cite{dormand1980family}, uses an adaptive step size. Recently, the Memory-efficient ALF Integrator~\cite{zhuang2021mali} guaranteeing constant memory cost shows good performance.
% \RED{Recently, based on the Asynchronous leapfrog (ALF) solver~\cite{mutze2013asynchronous}, the Memory-efficient ALF Integrator (MALI)~\cite{zhuang2021mali}, which guarantees constant memory cost regardless of the number of steps and has precise approximation, shows good performance.} 
However, when NODE learns a complex dataset, the step size of the ODE solver often becomes extremely small. Consequently, this results in dynamics equivalent to a substantial number of layers, thereby increasing the training time significantly. To address this issue, Jacobian and kinetic regularizations~\cite{finlay2020train} are introduced, simplifying the dynamics, increasing the step size, and reducing the training time.

The NODEs demonstrated superior performance in various tasks, including time series and others, by employing continuous modeling~\cite{jeon2021lightmove,jhin2021ace,hong2022prediction}. Specifically, the neural rough differential equations applied to the NODEs using rough-path theory showed good performance in the long-term time series task~\cite{morrill2021neuralrde,lee2022lord}. 

% In our model, we utilize the NODEs with various ODE solvers and regularizers.
In our model, we use a single linear layer as $f$ with various ODE solvers and regularizers to model times series dynamics effectively.
% In our model, we use the NODEs with various ODE solvers and regularizers to model times series dynamics effectively.

\subsection{Empirical Explorations on Decomposition and NODEs}

\begin{table}[t]
\setlength{\tabcolsep}{4pt}
\caption{MSE of forecasting results based on decomposition and NODE. We set the look-back window size as 336.}
\label{tab:mse_decom_node}
\begin{center}
\begin{tabular}{cccccc} \toprule
\textbf{Datasets}               & \begin{tabular}[c]{@{}c@{}}\textbf{Forecasting} \\ \textbf{horizon}\end{tabular} & \textbf{Linear} & \begin{tabular}[c]{@{}c@{}}\textbf{Linear}\\ \textbf{with T/R}\end{tabular} & \begin{tabular}[c]{@{}c@{}}\textbf{Linear}\\ \textbf{with T/S/R}\end{tabular} & \begin{tabular}[c]{@{}c@{}}\textbf{Linear}\\ \textbf{with NODE}\end{tabular} \\\midrule
\multirow{4}{*}{ETTh1} & 96                                                             & 0.375  & 0.375                                                     & 0.378                                                       & \textbf{0.371}                                             \\
                       & 192                                                            & 0.418  & 0.405                                                     & \textbf{0.404}                                              & 0.406                                                      \\
                       & 336                                                            & 0.479  & 0.439                                                     & \textbf{0.437}                                              & \textbf{0.437}                                             \\
                       & 720                                                            & 0.624  & \textbf{0.472}                                            & \textbf{0.472}                                              & 0.475                                                      \\\midrule
\multirow{4}{*}{ETTh2} & 96                                                             & 0.288  & 0.289                                                     & 0.288                                                       & \textbf{0.280}                                             \\
                       & 192                                                            & 0.377  & 0.383                                                     & \textbf{0.359}                                              & 0.364                                                      \\
                       & 336                                                            & 0.452  & 0.448                                                     & \textbf{0.422}                                              & 0.438                                                      \\
                       & 720                                                            & 0.698  & 0.605                                                     & 0.570                                                       & \textbf{0.568}                                 \\\bottomrule            
\end{tabular}
\end{center}
\end{table}

We conduct ablation studies to explore the efficacy of decomposition and NODE in LTSF. We implement simple variants of a single linear layer model by applying the decomposition method and NODE framework:
\begin{itemize}
    \item ``Linear'' is a single linear layer identical to the one introduced in~\cite{zeng2023transformers}. This model predicts future values based on past values via a weighted summation. The single linear layer is mathematically expressed as $\mathbf{\widehat{Y}=WX}$, where $\mathbf{W}\in\mathbb{R}^{P\times L}$ is a weight matrix.
    \item ``Linear with T/R'' decomposes time series into trend and residual components and then individually learns them using single linear layers, as proposed in~\cite{zeng2023transformers}.
    \item ``Linear with T/S/R'' decomposes time series into trend, seasonality, and residual components, and then individually learns them through single linear layers.
    %`Linear with NODE'' constructs the ODE function of the NODE framework using a single linear layer
    \item ``Linear with NODE'' also uses a single linear layer with the same structure as ``Linear'' to build the ODE function in the NODE framework. It employs the Euler method as the ODE solver.
    % ``Linear with NODE'' adds the NODE framework to ``Linear''. It uses a single linear layer to build the ODE function employs the Euler method as the ODE solver.
    
\end{itemize}   
% In a different representation from the expression above, 
The mathematical expression of ``Linear'' can also be depicted using the time variable $t$ as follows:
\begin{align}\label{eq:Linear}
    \boldsymbol{\widehat{x}}(t_1) = \boldsymbol{Wx}(t_0), 
\end{align}
where  $\boldsymbol{\widehat{x}}(t_1)$  is the future sequence $\mathbf{\widehat{Y}}$, $\boldsymbol{x}(t_0)$ is the historical sequence $\mathbf{X}$. $\boldsymbol{W}$ is a single linear layer matrix.
    
% Since the linear representation in Eq.~\eqref{eq:Linear} is too simple to capture the complex dynamics of time series. To overcome the limitation, we formulate the linear relationship as ODE to more accurately capture the dynamics of the time series. Our ODE formulation based on Eq.~\eqref{eq:Linear} is as follows:
``Linear with NODE'' formulation based on Eq.~\eqref{eq:Linear} is as follows:
\begin{align} 
        \frac{d\boldsymbol{x}(t)}{dt} &= \frac{1}{t_1-t_0}(\log_{}{\boldsymbol{W}})\boldsymbol{x}(t), \label{eq:ODE_base} \\ 
        \boldsymbol{x}(t_1) &= \boldsymbol{x}(t_0) + \int_{t_0}^{t_1} \frac{1}{t_1-t_0}(\log_{}{\boldsymbol{W}})\boldsymbol{x}(t) dt. \label{eq:integral_ODE_base}
\end{align}
The parameters are trained using the adjoint sensitivity method of the NODE framework.

Table~\ref{tab:mse_decom_node} shows the LTSF results for each variant model. Based on these observations, we infer the following findings:
\begin{enumerate}
    \item Applying time series decomposition methods enhances the performance of LTSF. This assertion is corroborated by the empirical observation that ``Linear with T/R'' shows better performance compared to the ``Linear''.
    \item In the ETTh2 dataset, ``Linear with T/S/R'' demonstrates a discernible advantage over ``Linear with T/R''. This suggests that the merits of a finer-grained decomposition, particularly the extraction of seasonality, might vary depending on the characteristics of the datasets.
    \item 
    % \BLUE{The ``Linear with NODE'' outperforms the ``Linear'', proving the benefits of utilizing the NODE framework in LTSF.}
    The better performance of ``Linear with NODE'' as compared to ``Linear'' demonstrates the benefits of incorporating the NODE framework in the domain of LTSF.
    % \item \RED{The utilization of the NODE framework in LTSF yields favorable outcomes, as the ``Linear with NODE'' outperforms the ``Linear''.}
\end{enumerate}

From these observations, we can infer that time-derivate modeling based on the NODE framework and more refined time series decomposition have a positive impact on LTSF. %granular

\begin{figure*}
    \begin{center}
        \includegraphics[width=0.9\textwidth]{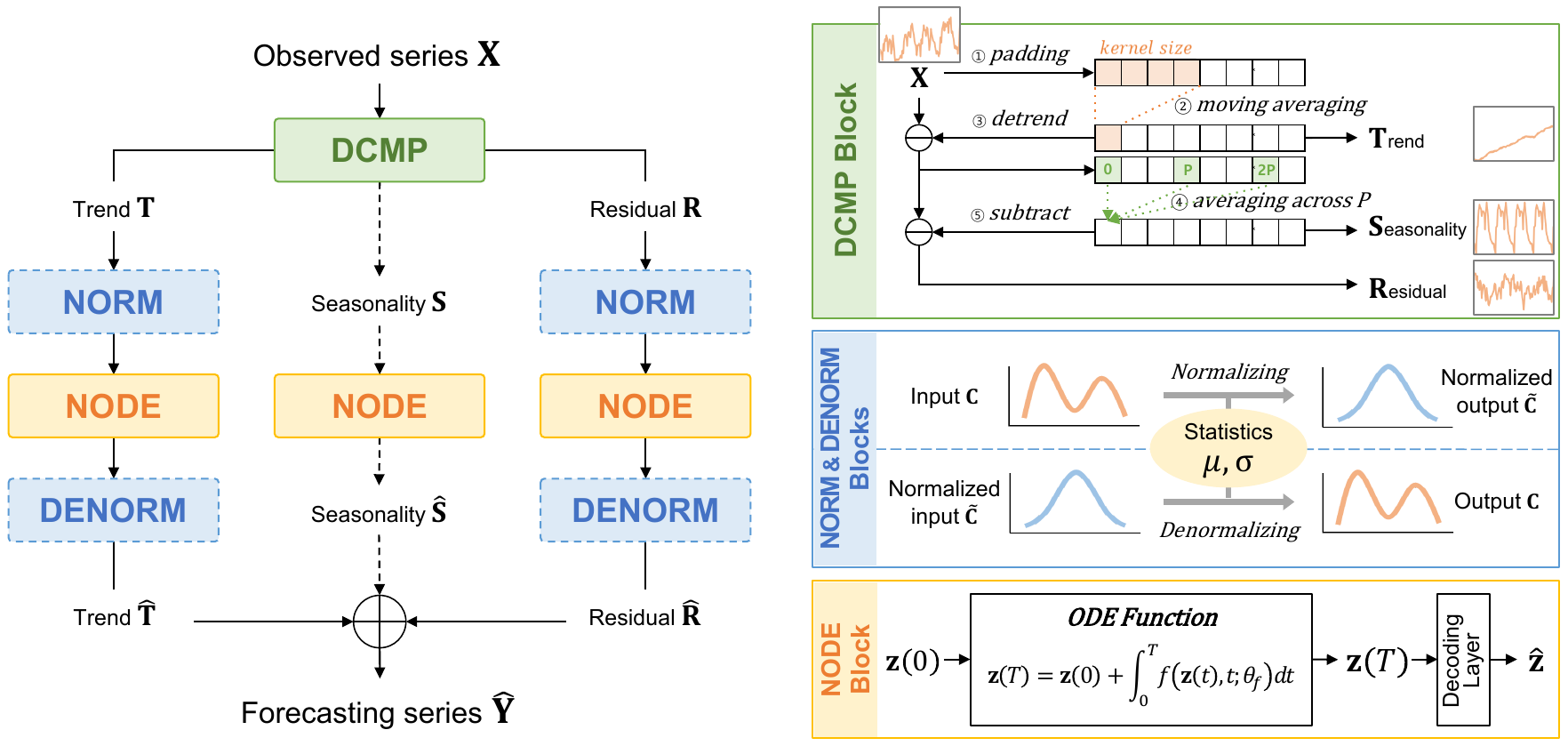}
    \end{center}
    \caption{The overall workflow of LTSF-DNODE. LTSF-DNODE consists of three blocks: the decomposition block (DCMP), the normalizing and denormalizing blocks (NORM \& DENORM), and the neural ODE block (NODE). DCMP block decomposes the observed time series into trend, seasonality, and residual components. NORM block normalizes the observed series based on its mean and variance. NODE block forecasts of the decomposed component based on each ODE. The processes indicated by dotted lines are optionally applied, considering the characteristics of the dataset.}
    \label{fig:overall_workflow}
\end{figure*}

\section{Proposed Method}
We explain the detailed information about our proposed model, LTSF-DNODE, in this section. We first describe an overview of how our model works, followed by detailed components and how they contribute to time series forecasting.

\subsection{Overall Workflow}
Fig.~\ref{fig:overall_workflow} illustrates the forecasting procedure of LTSF-DNODE. It consists of three main blocks: the decomposition block (DCMP), the normalizing and denormalizing blocks (NORM \& DENORM), and the NODE block (NODE). These blocks are sequentially applied to make predictions as follows:

\begin{enumerate}
    \item Find data characteristics with exploratory data analysis.
    \item An observed series $\mathbf{X}$ is given as input. The DCMP block decomposes $\mathbf{X}$ depending on data characteristics.
    % The DCMP block is used to decompose observed series $\mathbf{X}$ into three (or two) components.
    \item For datasets with distribution discrepancy problems, we apply the NORM block to address them.
    \item Then, the NODE block forecasts the future patterns of each decomposed component.
    \item In the case of the dataset normalized in (2), denormalization is performed in the DENORM block.
    \item Finally, the forecasting series $\widehat{\mathbf{Y}}$ is reconstructed by addition of the future decomposed components.
\end{enumerate}

\subsection{Exploratory Data Analysis}
% \subsection{Methods for Data Analyses}

\subsubsection{Properties}
To obtain insights from the datasets, we analyze their characteristics. The results of this analysis are presented in Table~\ref{tbl:data_info}. The fundamental structure and methodology of this analysis are derived from \cite{wang2022koopman}. 
The detailed methods used to acquire statistical properties are as follows:

\begin{itemize}
    \item ``Forecastability'' \cite{goerg2013forecastable} refers to a measure calculated by subtracting the entropy of the Fourier decomposition of the time series from one.
    \item ``Trend'' is the slope of the linear regression applied to the time series, adjusted according to its magnitude. 
    \item ``Seasonality'' is the ratio of noticeable patterns quantified by the ACF test \cite{witt1998testing}.
    \item ``Stationarity'' is measured by the ADF test on the residual component after removing trend and seasonality components from the time series. 
\end{itemize}

% The result and detailed explanation is in Experimental Evaluation.

\subsubsection{Decomposition}
DCMP block requires a pre-defined kernel size and seasonality period during the decomposition process. In order to achieve a suitable decomposition, we need to find the optimal parameters depending on the dataset. We consider various parameter combinations as candidates and conduct two tests, the auto-correlation function (ACF) test~\cite{witt1998testing} and the augmented Dickey-Fuller (ADF) test, to find the optimal values among them. 
% The ACF test examines whether a sequence has a seasonal pattern repeated at a determined cycle. The ADF test is applied to investigate the stationarity of the given sequence.
The ACF test examines if a sequence follows a repeating seasonal pattern with a pre-defined cycle, while the ADF test is used to check whether the sequence is stationary or not.

We first divide the time series into non-overlapping window-sized sequences and then decompose them. The ACF test is conducted for analyzing the seasonality components, while the ADF test is utilized to assess the residual components. Following the tests, we can quantify both seasonality and stationarity by calculating the proportion of sequences meeting the criteria relative to the total number of sequences.

This procedure is applied to all the datasets. 
% The experimental data-specific parameters and the results of exploratory analysis are summarized in Table~\ref{tbl:data_info}
An overview of the experimental parameters selected for each dataset and a summary of the results derived from the exploratory analysis are presented in Table~\ref{tbl:data_info}.
% The selected parameters for each dataset in the experiments and the results of the exploratory data analysis are summarized in Table~\ref{tbl:data_info}.%The final selected parameters for each dataset were summarized in Table~\ref{tbl:k_p_table}.
%\RED{Detailed parameter candidates and further analysis with these tests are specified in Experimental Evaluations.}

% Appendix} ~\ref{Additional Analysis}.

\subsubsection{Normalization}
We investigate the distribution of trend components in both training and testing data. If there exists a noticeable discrepancy between these distributions, we employ the NORM block to perform instance normalization. The same procedure is also applied to the residual components.
% We examine the training and testing data distributions of trends and residuals. If there is a noticeable distribution disparity issue observed in one or more features within the dataset, we employ the NORM block to perform instance normalization. 

The instance normalization method aids in mitigating distribution disparities and aligning the overall trend. Additional findings can be found in Section~\ref{sec:exp}. 

\subsection{DCMP Block} 

The DCMP block decomposes the time series into three, trend $\mathbf{T}$, seasonality $\mathbf{S}$, and residual $\mathbf{R}$ components, or two, without seasonality depending on the characteristics of the observed series $\mathbf{X}$. In the DCMP block, in order to extract the trend from the $\mathbf{X}$, we use the moving average method. To align the length of the derived trend with $\mathbf{X}$, we first apply $padding (\cdot)$, which involves pre-padding with the first value and post-padding with the last value of the input. After that, we perform an average pooling operation to extract the trend as follows: 
\begin{align}
    \mathbf{T} = AvgPool(padding(\mathbf{X})).
\end{align}

If the time series has a low significance of seasonality, extracting seasonality is omitted. In this case, the detrended ($\mathbf{X} - \mathbf{T}$) is regarded as the residual component, and the decomposition process is completed. However, if the series exhibits significant seasonality,  we extract the seasonality from $(\mathbf{X} - \mathbf{T})$. 
We first obtain seasonal fragments by averaging $(\mathbf{X} - \mathbf{T})$ across the pre-defined period $P$. In the fragments, each element is calculated as follows: 
\begin{align}\label{eq:seasonal}
    \mathbf{S}_{i} = {\frac{1}{m}} \sum_{k=0}^{m-1} (\mathbf{X}_{i+kP}-\mathbf{T}_{i+kP}),
\end{align} where $m$ is the smallest integer satisfying $L<i+mP$ for $0 \leq i < P$ and sequence length $L$, $\mathbf{S}_{i}$, $\mathbf{T}_{i+kP}$ and $\mathbf{X}_{i+kP}$ denote the $i$-th element of the seasonal fragments and the ($i+kP$)-th element of $\mathbf{T}$ and $\mathbf{X}$. 

We produce seasonal fragments with a period $P$ using Eq.~\eqref{eq:seasonal}. Then we construct the overall seasonality component $\mathbf{S}$ of length $L$ by tiling the obtained seasonal fragments. The residual component $\mathbf{R}$ is the remaining part of the observed series $\mathbf{X}$, obtained by subtracting the trend (and seasonality, depending on the dataset).

We effectively capture the temporal information from the time series using the decomposition method, which allows us to enhance the forecasting capabilities of our model. The analysis about the significance of seasonality for the datasets we use is in Section~\ref{sec:data_analyses}. 
% Section~\ref{sec:Analysis Decomposition}.

\subsection{NODE Block}
For the decomposition of the time series into trend, seasonality and residual, the NODEs can be written as follows and solved by the ODE solvers:
\begin{align} \label{eqn:ODE}
    \begin{split}
        \mathbf{T}(T) &= \mathbf{T}(0) + \int_{0}^{T} f_{\textbf{T}}(\mathbf{T}(t); \boldsymbol{\theta}_{\mathbf{T}}) dt,\\
        \mathbf{S}(T) &= \mathbf{S}(0) + \int_{0}^{T} f_\textbf{S}(\mathbf{S}(t) ; \boldsymbol{\theta}_{\mathbf{S}}) dt,\\ 
        \mathbf{R}(T) &= \mathbf{R}(0) + \int_{0}^{T} f_{\textbf{R}}(\mathbf{R}(t); \boldsymbol{\theta}_{\mathbf{R}}) dt,\\
    \end{split}
\end{align}
where the function $f_{\textbf{T}}: \mathbb{R}^L \rightarrow \mathbb{R}^L$ is a neural network with parameters $\boldsymbol{\theta}_{\mathbf{T}}$ that are learned from the data and captures the dynamics of the data with the trend. The function $f_{\textbf{T}}$ is defined as $f_{\textbf{T}}:= \frac{d\mathbf{T}(t)}{dt} = \boldsymbol{W}_{\textbf{T}}\mathbf{T}(t)$, where $\boldsymbol{W}_{\textbf{T}}\in\mathbb{R}^{L\times L}$ is a weight matrix associated with the trend component, modeled as a single linear layer. Starting with an initial value, $\mathbf{T}(0)=\mathbf{T}$, the NODE block computes the output, $\mathbf{T}(T)$, at the terminal time $T$ by solving the initial value problems. The $f_{\textbf{S}}$ and $f_{\textbf{R}}$ are defined and parameterized in the same manner as $f_{\textbf{T}}$. 
% The functions $f_{\textbf{S}}$ and $f_{\textbf{R}}$ are defined and parameterized in the same manner as $f_{\textbf{T}}$. 

Note that we use various ODE solvers with the Jacobian and kinetic regularization~\cite{finlay2020train}. Leveraging these regularizers allows for more accurate regularization of the learned dynamics at each time point, which exhibits diverse patterns. 
Since NODE requires the same dimension size of input and output, the results of the ODE solvers are decoded into the forecasting horizon through the decoding layer, as follows:
\begin{align} \label{eqn:forcast}
    \begin{split}
        \mathbf{\widehat{T}} &= {\texttt{FC}^{\mathbf{T}}}_{L \rightarrow H} (\mathbf{T}(T)) , \\
        \mathbf{\widehat{S}} &= {\texttt{FC}^{\mathbf{S}}}_{L \rightarrow H} (\mathbf{S}(T)) , \\ 
        \mathbf{\widehat{R}} &= {\texttt{FC}^{\mathbf{R}}}_{L \rightarrow H} (\mathbf{R}(T)) , \\
    \end{split}
\end{align}
where each ${\texttt{FC}}_{L \rightarrow H}$ means a fully-connected layer whose input size is $L$ and output size is $H$.
% \RED{where each ${\texttt{FC}^i}_{L \rightarrow H}$ means a fully-connected layer whose input size is $L$ and output size is also $H$. }

% The values obtained through integration correspond to the values at \RED{time $T$} with respect to the integration variable $t$. In other words, they represent the target sequence, which is the sequence after the observed sequence. Ultimately, 
The decomposed components $\mathbf{T}$, $\mathbf{S}$, and $\mathbf{R}$ of the observed sequence yield the future decomposed components $ \mathbf{\widehat{T}}$, $ \mathbf{\widehat{S}}$, and $\mathbf{\widehat{R}}$ after passing through the NODE block.

\subsection{NORM \& DENORM Blocks}
If the dataset has distribution discrepancy problems, we apply instance normalization to the trend and residual components using the NORM and DENORM blocks. In Section~\ref{sec:data_analyses}, it will be discussed why instance normalization is only done on the trend and residual components. Instance normalization is performed on feature dimensions. 
Specifically, given the original component $\mathbf{C}$, which could be a trend or a residual, represented as $\mathbf{C}_{ij} \in \mathbb{R}^{{L} \times {F}}$, the normalization procedure can be expressed as follows:
\begin{align}
    \boldsymbol{\mu}_{i} = \frac{1}{F} \sum_{j=1}^{F} \mathbf{C}_{ij} \,&, \quad
    {\boldsymbol{\sigma}_i}^2 = \frac{1}{F} \sum_{j=1}^{F} (\mathbf{C}_{ij}-\boldsymbol{\mu}_i )^2 , \\
    \mathbf{\widetilde{C}}_{ij} &= \frac{\mathbf{C}_{ij} - \boldsymbol{\mu}_{i}}{\boldsymbol{\sigma}_i},
\end{align} where $L$ is the length of the sequence, $F$ is the number of features, $\boldsymbol{\mu}_{i}$ and ${\boldsymbol{\sigma}_i}^2$ are the $i$-th mean and variance of the original data, and $\mathbf{\widetilde{C}}_{ij}$ is normalized component. 
%In the case of a normalized dataset, trend and residual are denormalized through the DENORM block before being reconstructed into a forecasting series $\mathbf{\widehat{Y}}$. Once component NODE block has made its output, denote it \RED{${\mathbf{\widetilde{D}}}_{ij}$}, it goes through the DENORM block. 

DENORM block conducts the inverse process of normalization using the preserved mean ($\boldsymbol{\mu}_i$) and standard deviation ($\boldsymbol{\sigma}_i$) obtained during the normalization process. When normalized component $\widetilde{\mathbf{C}}$ is given, the denormalization procedure performs the reverse of normalization as follows:
\begin{align}
    \mathbf{C}_{ij} = \sigma_i{\mathbf{\widetilde{C}}_{ij} + \boldsymbol{\mu}_{i}},
\end{align}
% In place of $\mathbf{D}$, trend $\mathbf{T}$ or residual $\mathbf{R}$ can be substituted.

\subsection{Forecasting}
To forecast future series, if the dataset is normalized, we apply denormalization to restore the original distribution, and we add up $\mathbf{\widehat{T}}$, $\mathbf{\widehat{S}}$, and $\mathbf{\widehat{R}}$ to obtain the final prediction $\mathbf{\widehat{Y}}$:
\begin{align} \label{eqn:forecasting}
    \begin{split}
        \mathbf{\widehat{Y}} = \mathbf{\widehat{T}} + \mathbf{\widehat{S}} + \mathbf{\widehat{R}}.
    \end{split}
\end{align}
The values used as Trend regularizers are defined as follows:
\begin{align}
    \dot{E}_\mathbf{T}(t)&={||f_\mathbf{T}(\mathbf{T}(t);\theta)||}^2,  \\
    \dot{J}_\mathbf{T}(t)&={||{\epsilon}^\top{\nabla}f_\mathbf{T}(\mathbf{T}(t);\theta)||},
\end{align}where `$\cdot$' denotes derivative, $\epsilon$ is sampled from standard normal distribution, and $f_\mathbf{T}$ refers to the function defined in Eq.~\eqref{eqn:ODE}. The same definitions are applied to the seasonality and residual components as well.

The proposed model is trained using the mean square error (MSE) loss function with regularization terms as follows:
\begin{align}\label{eqn:loss_mse}
    \begin{split}
        Loss &= \frac{\sum_{i=1}^{n} (\mathbf{Y}_{i}-\mathbf{\widehat{Y}}_{i})^{2}}{n} \\
        &+\sum_{j\in\{\mathbf{T}, \mathbf{S}, \mathbf{R}\}}(\lambda_K{E_j(t)} + \lambda_J{J}_j(t))
    \end{split}    
\end{align} where $\mathbf{Y}_i$ is the $i$-th element of the ground truth $\mathbf{Y}$, $n$ is the number of elements, and $\{\mathbf{T}, \mathbf{S}, \mathbf{R}\}$ each represent the Trend, Seasonality, and Residual in the summation, respectively. The coefficients $\lambda_K$ and $\lambda_J$ are used as hyperparameters.

\subsection{Comparison with Existing Models}

\begin{table}[]
    \centering
    \setlength{\tabcolsep}{0.5pt}
    \caption{Comparison with existing baseline models, which use decomposition and normalization}
    \label{tab:comparison table}
    \begin{tabular}{cccc} \toprule
    Models                      & Main Architecture           & Decomposition                                                                & Normalization                                                                     \\\midrule
    FEDformer~\cite{zhou2022fedformer}                   & Transformer                 & Frequency                                                                    & -                                                                                 \\
    Autoformer~\cite{wu2021autoformer}                  & Transformer                 & Fixed (T/S)                                                                  & -                                                                                 \\
    NLinear~\cite{zeng2023transformers}                     & Single Linear               & -                                                                            & Subtraction                                                                       \\
    DLinear~\cite{zeng2023transformers}                     & Single Linear               & Fixed (T/R)                                                                  & -                                                                                 \\\midrule
    \multirow{2}{*}{LTSF-DNODE} & \multirow{2}{*}{Neural ODE} & \multirow{2}{*}{\begin{tabular}[c]{@{}c@{}}Selective\\ (T/S/R)\end{tabular}} & \multirow{2}{*}{\begin{tabular}[c]{@{}c@{}}Instance\\ Normalization\end{tabular}} \\
                                &                             &                                                                              &                                                                                  \\\bottomrule
    \end{tabular}
\end{table}

\begin{table*}[]
\begin{center}
\caption{MSE and MAE results for multivariate time series forecasting on benchmark datasets. We set the look-back window size as 104 for ILI and 336 for the others. The ILI dataset has a forecasting horizon $H \in \{24, 36, 48, 60\}$. For the others, $H \in \{96, 192, 336, 720\}$. The best results are highlighted in \textbf{bold}.}
\label{tbl:main_table}
\resizebox{1.0\textwidth}{!}{
\begin{tabular}{cccccccccccccccccc} \toprule
\multirow{2}{*}{\textbf{Datasets}}    & \multirow{2}{*}{\begin{tabular}[c]{@{}c@{}}\textbf{Forecasting}\\ \textbf{horizon}\end{tabular}} & \multicolumn{2}{c}{\textbf{LTSF-DNODE}}  & \multicolumn{2}{c}{\textbf{NLinear}}     & \multicolumn{2}{c}{\textbf{DLinear}}     & \multicolumn{2}{c}{\textbf{FEDformer}} & \multicolumn{2}{c}{\textbf{Autoformer}} & \multicolumn{2}{c}{\textbf{Informer}} & \multicolumn{2}{c}{\textbf{Pyraformer}} & \multicolumn{2}{c}{\textbf{LogTrans}} \\\cmidrule(lr){3-4} \cmidrule(lr){5-6} \cmidrule(lr){7-8} \cmidrule(lr){9-10} \cmidrule(lr){11-12} \cmidrule(lr){13-14} \cmidrule(lr){15-16} \cmidrule(lr){17-18}
                             &                                                                                & \textbf{MSE}            & \textbf{MAE}            & \textbf{MSE}            & \textbf{MAE}            & \textbf{MSE}            & \textbf{MAE}            & \textbf{MSE}           & \textbf{MAE}           & \textbf{MSE}            & \textbf{MAE}           & \textbf{MSE}           & \textbf{MAE}          & \textbf{MSE}            & \textbf{MAE}           & \textbf{MSE}           & \textbf{MAE}          \\\midrule
\multirow{4}{*}{Electricity} & 96                                                                             & \textbf{0.140} & \textbf{0.237} & 0.141          & \textbf{0.237} & \textbf{0.140} & \textbf{0.237} & 0.193         & 0.308         & 0.201          & 0.317         & 0.274         & 0.368        & 0.386          & 0.449         & 0.258         & 0.357        \\
                             & 192                                                                            & \textbf{0.153} & 0.249 & 0.154          & \textbf{0.248} & \textbf{0.153} & 0.249          & 0.201         & 0.315         & 0.222          & 0.334         & 0.296         & 0.386        & 0.386          & 0.443         & 0.266         & 0.368        \\
                             & 336                                                                            & \textbf{0.168} & 0.267          & 0.171          & \textbf{0.265} & 0.169          & 0.267          & 0.214         & 0.329         & 0.231          & 0.338         & 0.300         & 0.394        & 0.378          & 0.443         & 0.280         & 0.380        \\
                             & 720                                                                            & \textbf{0.203} & 0.300          & 0.210          & \textbf{0.297} & \textbf{0.203} & 0.301          & 0.246         & 0.355         & 0.254          & 0.361         & 0.373         & 0.439        & 0.376          & 0.445         & 0.283         & 0.376        \\\midrule
\multirow{4}{*}{Exchange}    & 96                                                                             & \textbf{0.078} & \textbf{0.200} & 0.089          & 0.208          & 0.081          & 0.203          & 0.148         & 0.278         & 0.197          & 0.323         & 0.847         & 0.752        & 0.376          & 1.105         & 0.968         & 0.812        \\
                             & 192                                                                            & \textbf{0.155} & \textbf{0.292} & 0.180          & 0.300          & 0.157          & 0.293          & 0.271         & 0.380         & 0.300          & 0.369         & 1.204         & 0.895        & 1.748          & 1.151         & 1.040         & 0.851        \\
                             & 336                                                                            & \textbf{0.259} & \textbf{0.388} & 0.331          & 0.415          & 0.305          & 0.414          & 0.460         & 0.500         & 0.509          & 0.524         & 1.672         & 1.036        & 1.874          & 1.172         & 1.659         & 1.081        \\
                             & 720                                                                            & \textbf{0.606} & \textbf{0.591} & 1.033          & 0.780          & 0.643          & 0.601          & 1.195         & 0.841         & 1.447          & 0.941         & 2.478         & 1.310        & 1.943          & 1.206         & 1.941         & 1.127        \\\midrule
\multirow{4}{*}{Weather}     & 96                                                                             & \textbf{0.174} & 0.234          & 0.182          & \textbf{0.232} & 0.176          & 0.237          & 0.217         & 0.296         & 0.266          & 0.336         & 0.300         & 0.384        & 0.896          & 0.556         & 0.458         & 0.490        \\
                             & 192                                                                            & \textbf{0.215} & 0.272          & 0.225          & \textbf{0.269} & 0.220          & 0.282          & 0.276         & 0.336         & 0.307          & 0.367         & 0.598         & 0.544        & 0.622          & 0.624         & 0.658         & 0.589        \\
                             & 336                                                                            & \textbf{0.262} & 0.311          & 0.271          & \textbf{0.301} & 0.265          & 0.319          & 0.339         & 0.380         & 0.359          & 0.395         & 0.578         & 0.523        & 0.739          & 0.753         & 0.797         & 0.652        \\
                             & 720                                                                            & \textbf{0.323} & 0.362          & 0.338          & \textbf{0.348} & \textbf{0.323} & 0.362          & 0.403         & 0.428         & 0.419          & 0.428         & 1.059         & 0.741        & 1.004          & 0.934         & 0.869         & 0.675        \\\midrule
\multirow{4}{*}{ILI}         & 24                                                                             & \textbf{1.626} & \textbf{0.848}          & 1.683          & 0.858 & 2.215          & 1.081          & 3.228         & 1.260         & 3.483          & 1.287         & 5.764         & 1.677        & 1.420          & 2.012         & 4.480         & 1.444        \\
                             & 36                                                                             & \textbf{1.589} & \textbf{0.845}          & 1.703          & 0.859 & 1.963          & 0.963          & 2.679         & 1.080         & 3.103          & 1.148         & 4.755         & 1.467        & 7.394          & 2.031         & 4.799         & 1.467        \\
                             & 48                                                                             & \textbf{1.566}          & \textbf{0.861}          & 1.719 & 0.884 & 2.130          & 1.024          & 2.622         & 1.078         & 2.669          & 1.085         & 4.763         & 1.469        & 7.551          & 2.057         & 4.800         & 1.468        \\
                             & 60                                                                             & \textbf{1.693} & \textbf{0.911} & 1.819          & 0.917          & 2.368          & 1.096          & 2.857         & 1.157         & 2.770          & 1.125         & 5.264         & 1.564        & 7.662          & 2.100         & 5.278         & 1.560        \\\midrule
\multirow{4}{*}{ETTh1}       & 96                                                                             & \textbf{0.369} & \textbf{0.392} & 0.374          & 0.394          & 0.375          & 0.399          & 0.376         & 0.419         & 0.449          & 0.459         & 0.865         & 0.713        & 0.664          & 0.612         & 0.878         & 0.740        \\
                             & 192                                                                            & \textbf{0.403} & \textbf{0.411} & 0.408          & 0.415          & 0.405          & 0.416          & 0.420         & 0.448         & 0.500          & 0.482         & 1.008         & 0.792        & 0.790          & 0.681         & 1.037         & 0.824        \\
                             & 336                                                                            & \textbf{0.423} & \textbf{0.422} & 0.429          & 0.427          & 0.439          & 0.443          & 0.459         & 0.465         & 0.521          & 0.496         & 1.107         & 0.809        & 0.891          & 0.738         & 1.238         & 0.932        \\
                             & 720                                                                            & \textbf{0.425} & \textbf{0.444} & 0.440          & 0.453          & 0.472          & 0.490          & 0.506         & 0.507         & 0.514          & 0.512         & 1.181         & 0.865        & 0.963          & 0.782         & 1.135         & 0.852        \\\midrule
\multirow{4}{*}{ETTh2}       & 96                                                                             & \textbf{0.272} & \textbf{0.334} & 0.277          & 0.338          & 0.289          & 0.353          & 0.346         & 0.388         & 0.358          & 0.397         & 3.755         & 1.525        & 0.645          & 0.597         & 2.116         & 1.197        \\
                             & 192                                                                            & \textbf{0.334} & \textbf{0.375} & 0.344          & 0.381          & 0.383          & 0.418          & 0.429         & 0.439         & 0.456          & 0.452         & 5.602         & 1.931        & 0.788          & 0.683         & 4.315         & 1.635        \\
                             & 336                                                                            & \textbf{0.341} & \textbf{0.390} & 0.357          & 0.400          & 0.448          & 0.465          & 0.496         & 0.487         & 0.482          & 0.486         & 4.721         & 1.835        & 0.907          & 0.747         & 1.124         & 1.604        \\
                             & 720                                                                            & \textbf{0.387} & \textbf{0.428} & 0.394          & 0.436          & 0.605          & 0.551          & 0.463         & 0.474         & 0.515          & 0.511         & 3.647         & 1.625        & 0.963          & 0.783         & 3.188         & 1.540        \\\midrule
\multirow{4}{*}{ETTm1}       & 96                                                                             & \textbf{0.299} & \textbf{0.343} & 0.306          & 0.348          & \textbf{0.299} & \textbf{0.343} & 0.379         & 0.419         & 0.505          & 0.475         & 0.672         & 0.571        & 0.543          & 0.510         & 0.600         & 0.546        \\
                             & 192                                                                            & \textbf{0.334} & \textbf{0.364} & 0.349          & 0.375          & 0.335          & 0.365          & 0.426         & 0.441         & 0.553          & 0.496         & 0.795         & 0.669        & 0.557          & 0.537         & 0.837         & 0.700        \\
                             & 336                                                                            & \textbf{0.369} & \textbf{0.386} & 0.375          & 0.388          & \textbf{0.369} & \textbf{0.386} & 0.445         & 0.459         & 0.621          & 0.537         & 1.212         & 0.871        & 0.754          & 0.655         & 1.124         & 0.832        \\
                             & 720                                                                            & \textbf{0.424} & \textbf{0.419} & 0.433          & 0.422          & 0.425          & 0.421          & 0.543         & 0.490         & 0.671          & 0.561         & 1.166         & 0.823        & 0.908          & 0.724         & 1.153         & 0.820        \\\midrule
\multirow{4}{*}{ETTm2}       & 96                                                                             & \textbf{0.163} & \textbf{0.253} & 0.167          & 0.255          & 0.167          & 0.260          & 0.203         & 0.287         & 0.255          & 0.339         & 0.365         & 0.453        & 0.435          & 0.507         & 0.768         & 0.642        \\
                             & 192                                                                            & \textbf{0.217} & \textbf{0.291} & 0.221          & 0.293          & 0.224          & 0.303          & 0.269         & 0.328         & 0.281          & 0.340         & 0.533         & 0.563        & 0.730          & 0.673         & 0.989         & 0.757        \\
                             & 336                                                                            & \textbf{0.270} & \textbf{0.325} & 0.274          & 0.327          & 0.281          & 0.342          & 0.325         & 0.366         & 0.339          & 0.372         & 1.363         & 0.887        & 1.201          & 0.845         & 1.334         & 0.872        \\
                             & 720                                                                            & \textbf{0.359} & \textbf{0.381} & 0.368          & 0.384          & 0.397          & 0.421          & 0.421         & 0.415         & 0.433          & 0.432         & 3.379         & 1.338        & 3.625          & 1.451         & 3.048         & 1.328        \\\bottomrule
\end{tabular}}
\end{center}
\end{table*}

%Informer   & Transformer   & - & - \\
%Pyraformer & Transformer   & - & - \\
%LongTrans  & Transformer   & - & - \\

% \RED{\TODO: 설명 보충 - 우리 모델에서 NODE block이 빠지면 NLinear 또는 DLinear와 비슷해질 수 있음}
% , NODE block이 없으면 sub-optimal, underfitting 해짐 등 $\rightarrow$ sub-optimal같은 것을 말하려면, curve를 보여줘야함. 아래 문단에 추가안해도될것같습니다.
% This subsection includes a comparison of our model to other baseline models, which use decomposition and normalization. In contrast to the baseline models using single linear and Transformer, our model employs NODE as the main architecture. Furthermore, we utilize both \textcolor{blue}{normalization and decomposition that differ from those utilized by other baselines. $\rightarrow$ 
Table~\ref{tab:comparison table} shows the main differences between our model and the baselines. In contrast to the baseline models using single linear and Transformer, our model employs NODE as the main architecture. If we remove the NODE framework from our proposed model, it closely resembles a combination of the NLinear and DLinear. However, our approach employs both data characteristic-dependent decomposition and normalization methods, as opposed to the consistent method used in other baselines. The normalization and decomposition methods of each model can be found in Table~\ref{tab:comparison table}.

%Both NLinear and DLinear employ a single linearlayer. On the other hand, our model incorporates the single linear layer into a Neural ODE. As a result, we enable learning from a time derivative perspective while also allowing the use of various additional structures like regularizers. FEDformer and Autoformer also have the decomposition methods, but these decomposition methods are applied uniformly to all datasets, disregarding the unique characteristics of the datasets. By contrast, we use decomposition based on the results of statistical analysis specific to the datasets.

\section{Experimental Evaluations} \label{sec:exp}

In this section, we compare the performance of LTSF-DNODE for multivariate LTSF on real-world datasets against baselines and provide an analysis of the model's architecture with respect to data characteristics.

\subsection{Experimental Environments}
All experiments are conducted in the following software and hardware environments: \textsc{Ubuntu} 18.04.4 LTS, \textsc{Python} 3.8.13, \textsc{Pytorch} 1.11.0, \textsc{CUDA} V10.0.130, \textsc{NVIDIA Quadro RTX 6000/8000}.

We select several cases for each dataset by varying hyper-parameters. The learning rates are in \{0.05, 0.01, 0.005, 0.001, 0.0005, 0.0001\} and the batch sizes are in \{8, 16, 32, 64\}. The number of training epochs is set to a maximum of 100. With the validation dataset, an early-stop approach with a patience of 10 iterations is applied. We implement the ODE function of the NODE framework using a single linear layer without an activation function and use Euler, RK4, and DOPRI as the ODE solvers. The coefficients of the Jacobian and kinetic regularizers are in \{0.1, 0.2, \ldots, 1.0\}. The terminal time, denoted as $T$, is set to 1.

% Additional hyperparameter settings are in Appendix ~\ref{Hyperparameters}.

\begin{table*} []
    \small
    \centering
    \setlength{\tabcolsep}{2.5pt}
    \caption{The analysis results of the benchmark datasets. The selected kernel size and period yields the highest seasonality.}
    \label{tbl:data_info}
    \begin{tabular}{cccccccccc} \toprule
        \textbf{Datasets}                             & \textbf{Electricity} & \textbf{Exchange}  & \textbf{Weather} & \textbf{ILI}      & \textbf{ETTh1}     & \textbf{ETTh2}     & \textbf{ETTm1}     & \textbf{ETTm2}      \\ \midrule
        \multicolumn{1}{c}{Features}        & 321         & 8        & 21       & 7        & 7         & 7         & 7         & 7          \\
        \multicolumn{1}{c}{Timesteps}       & 26,304       & 7,588    & 52,696     & 966      & 17,420     & 17,420     & 69,680     & 69,680   \\
        \multicolumn{1}{c}{Granularity}     & 1hour          & 1day        & 10min      & 1week       & 1hour        & 1hour        & 15min       & 15min   \\
        \multicolumn{1}{c}{Forecastability} & 0.126       & 0.159       & 0.141     & 0.173    & 0.148     & 0.156     & 0.142     & 0.144      \\
        \multicolumn{1}{c}{Trend}           & \num{-4.00E-06}   & \num{1.84E-04} & \num{4.00E-06}  & \num{1.46E-03} & \num{-1.90E-05} & \num{-7.90E-05} & \num{-5.40E-05} & \num{-2.00E-04}\\
        \multicolumn{1}{c}{Kernel size} & 25  & 10 &10 &25 & 10    & 25    & 50    & 25    \\
        \multicolumn{1}{c}{Period}  & 24 & 7  & 6        & 52  & 48    & 24    & 7     & 7     \\
        \multicolumn{1}{c}{Seasonality}     & 98.70\%     & 11.25\%   & 52.38\%    & 22.22\%  & 97.02\%   & 93.45\%   & 94.20\%   & 80.21\%  \\
        \multicolumn{1}{c}{Stationarity}    & 100.00\%    & 100.00\% & 99.87\%   & 96.83\%  & 100.00\%  & 100.00\%  & 100.00\%  & 100.00\%\\ \bottomrule
        \end{tabular}
\end{table*}

\subsubsection{Datasets}
In order to evaluate LTSF-DNODE, we conduct experiments on real-world datasets with diverse characteristics collected from various domains such as energy, economics, and more. These datasets possess a wide range of features (see Table~\ref{tbl:data_info} for its detailed descriptions).
\begin{itemize}
    \item Electricity
    % \footnote{https://archive.ics.uci.edu/ml/datasets/ElectricityLoadDiagrams20112014}
    records the amount of electricity consumption of 321 customers from 2012 to 2014.
    \item Exchange Rate~\cite{lai2018modeling}
    % \footnote{https://github.com/laiguokun/multivariate-time-series-data}
    consists of the daily records of the exchange rates of eight countries from 1990 to 2016.
    \item Weather
    % \footnote{https://www.bgc-jena.mpg.de/wetter/}
    consists of data from 21 weather-related features (e.g. air temperature, humidity), recorded in Germany with 10-min interval during 2020.
    \item ILI (Influenza-like Illness)
    % \footnote{https://gis.cdc.gov/grasp/fluview/fluportaldashboard.html}
    is compiled by the Centers for Disease Control and Prevention of the United States from 2002 to 2021.
    \item ETT (Electricity Transformer Temperature)~\cite{zhou2021informer}
    % \footnote{https://github.com/zhouhaoyi/ETDataset} 
    consists of two datasets with hourly granularity and two datasets with 15-minute granularity. Each data shows seven oil and load related properties of transformer. The data was aggregated from July 2016 to July 2018.
\end{itemize}

\subsubsection{Baselines}
We consider the following 7 baselines for long-term time series forecasting. These baselines consist of five Transformer-based models and two Linear-based models, the previous state-of-the-art models:
\begin{itemize}
    \item Linear-based models: NLinear and DLinear~\cite{zeng2023transformers}
    \item Transformer-based models: Informer~\cite{zhou2021informer}, Pyraformer~\cite{liu2021pyraformer}, and LogTrans~\cite{li2019enhancing}
    \item Tranformer-based models with decomposition method: FEDformer~\cite{zhou2022fedformer}, and Autoformer~\cite{wu2021autoformer}
\end{itemize}

\subsubsection{Metrics}

To evaluate our model, we use MSE and mean absolute error (MAE). These metrics are as follows:
\begin{align}\label{eqn:metric}
    \textrm{MSE} = \frac{\sum_{i=1}^{n} (\mathbf{Y}_{i}-\mathbf{\widehat{Y}}_{i})^{2}}{n},\;
    \textrm{MAE} = \frac{1}{n} \sum_{i=1}^{n}\Bigl\lvert\frac{\mathbf{Y}_{i}-\mathbf{\widehat{\mathbf{Y}}}_{i}}{\mathbf{Y}_{i}}\Bigr\rvert,
\end{align} where $\mathbf{Y}_i$ is the $i$-th element of the ground truth $\mathbf{Y}$, $n$ is the number of elements.

\subsection{Main Results}
In Table~\ref{tbl:main_table}, we list the results of the multivariate long-term time series forecasting. The forecasting horizon refers to the length of the sequence that the model aims to predict.
%, while improvement (Improve) measures how much better the LTSF-DNODE model performs compared to the previous state-of-the-art model baseline. 
LTSF-DNODE shows better performance in most cases than the previous models.
% \RED{Compared to DLinear, LTSF-DNODE achieved an average improvement of \RED{1.84\%} on MSE and \RED{0.43\%} on MAE.}
This result indicates that LTSF-DNODE effectively utilized the essential temporal information emphasized in \cite{zeng2023transformers} for LTSF, leveraging precise preprocessing methods and forecasting modeling based on NODE.

There are performance improvements (e.g., up to 15.08\% on MSE and up to 6.28\% on MAE) observed in the Exchange, ILI, ETTh1, ETTh2, and ETTm2 datasets. 
% \RED{These improvements are due to the strong seasonality in these datasets as well as the distribution discrepancy between the train and test datasets.}
In the Electricity, Weather, and ETTm1 datasets, our model demonstrated performance similar to that of Linear-based models.

These improvements are attributed to decomposition and normalization methods tailored to the data characteristics. In addition, it can be argued that formulating between past and future values as Eq.~\eqref{eq:ODE_base} contributed positively to LTSF by accurately capturing the dynamics of time series. The subsequent section discusses the analysis results of data characteristics and how each component of our proposed model contributed to performance enhancement. 

\begin{figure}[t]
    \begin{center}
        \subfigure[ILI OT feature]{\includegraphics[width=0.49\textwidth]{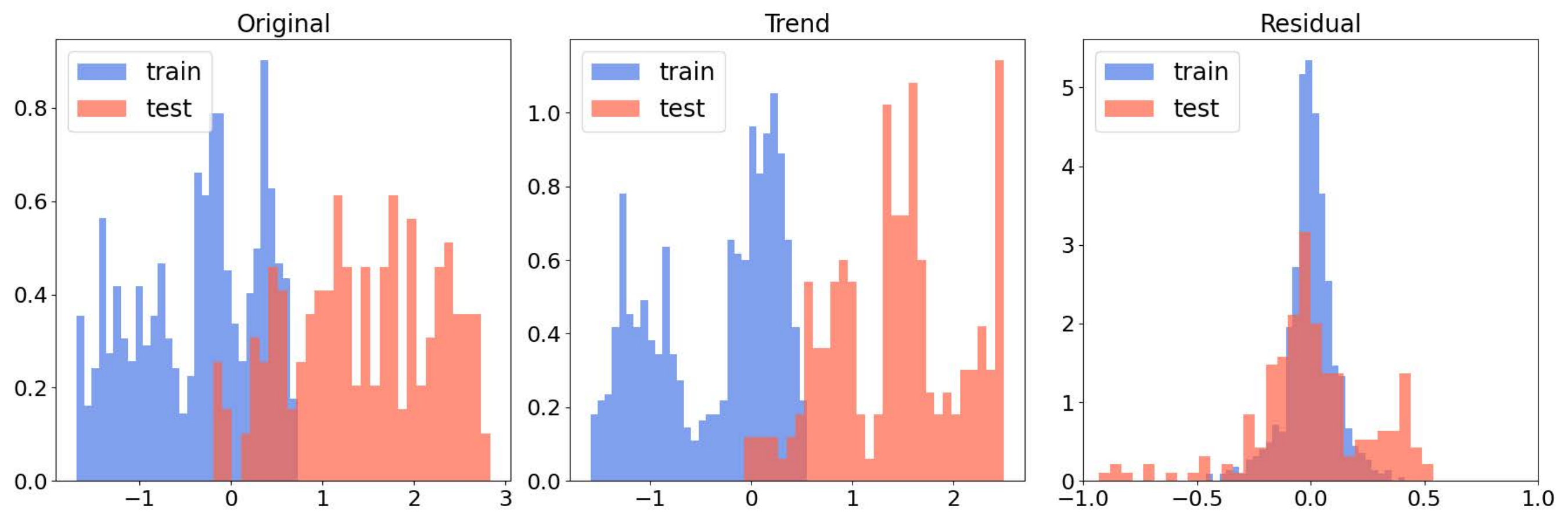}}
        \subfigure[ETTh1 OT feature]{\includegraphics[width=0.49\textwidth]{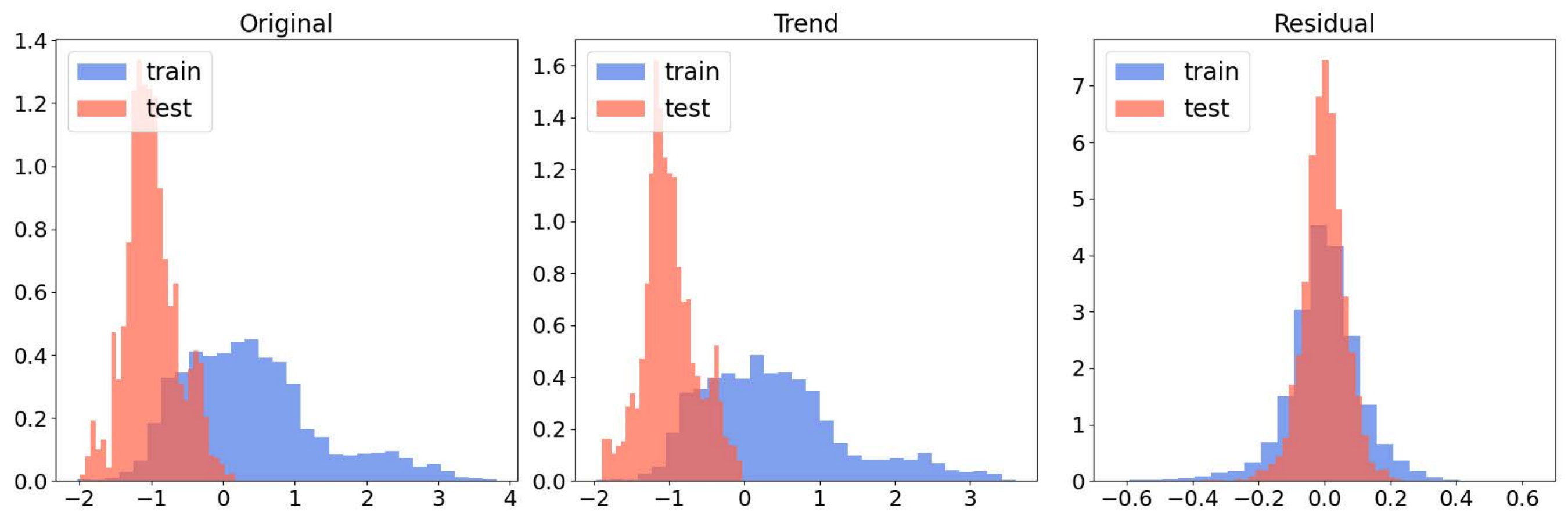}}
        % \subfigure[ETTh2 HULL feature]{\includegraphics[width=0.49\textwidth]{figures/ETTh2_feature_2.pdf}}
        % \subfigure[ETTm2 OT feature]{\includegraphics[width=0.49\textwidth]{figures/ETTm2_feature_7.pdf}}
    \end{center}
    % \caption{Distribution of original, trend, and residual for feature OT in (a) ILI, (b) ETTh1, (d) ETTm2, and feature HULL in (c) ETTh2  datasets. There is a clear discrepancy in the distribution between the train and test datasets.}
    \caption{Distribution of original, trend, and residual for feature OT in (a) ILI and (b) ETTh1. There is a clear discrepancy in the distribution between the train and test datasets.}
    \label{fig:dist_disc}
\end{figure}

\subsection{Data Analysis Results} \label{sec:data_analyses}

% \subsubsection{Decomposition} \label{sec:Analysis Decomposition}
% 여기서 각각 결정된 period kernelsize를 위의 표 결과와 비교하며 해석. 분해법을 사용해야하는 이유를 보여준다면 적절하지 못한 다른 커널사이즈로 모델을 적합시킬 경우 성능이 저하됨을 보여줄 수 있을 듯.

% DCMP block requires a predetermined kernel size and seasonality period during the decomposition process. In order to achieve a proper decomposition, we needed to find the optimal parameters depending on the data. We considered various parameter combinations as candidates and conducted two tests, the auto-correlation function (ACF) test~\cite{witt1998testing} and the augmented Dickey-Fuller (ADF) test, to find the optimal values among them. The ACF test examines whether a sequence has a seasonal pattern repeated at a determined cycle. The ADF test is applied to investigate the stationarity of the given sequence.
% \RED{Detailed parameter candidates and further analysis with these tests are specified in Appendix} ~\ref{Additional Analysis}.

% We first divide the dataset into non-overlapping window-sized sequences and then decomposed them. The ACF test is conducted for analyzing the seasonality components, while the ADF test is conducted for analyzing the residual components. After tests, we can quantify seasonality and stationarity based on the ratio of sequences that satisfy the criteria compared to the total number of sequences.
The result of exploratory analysis on each dataset can be found in Table~\ref{tbl:data_info}. ``Forecastability'' and ``Trend'' metrics are evaluated for the full sequence, with averaging across the features. ``Seasonality'' is determined based on sub-sequences with a length (104 for ILI and 336 for the others) using kernel sizes of \{10, 25, 50\} and periods that are determined based on the ``Granularity'' (e.g., the period is 24 for an hour-based dataset to determine the seasonality of a day, 48 for two days). In Table~\ref{tbl:data_info}, ``Seasonality'' is the highest ratio obtained among combinations of kernel sizes and periods. Similarly, we calculate the proportion of sequences exhibiting stationarity for each feature and take the average sequences sliced of length 720 (also 104 for ILI).

Based on the results of the ACF test, we select three candidates with the highest seasonality ratio for each kernel size. Then, using the outcomes of the ADF test, we compare the stationarity ratio of the residuals to identify the best parameter combination. Since the residual should show stationarity, we select the parameter combination that shows a higher ratio of stationarity in the residual. 
If it is ambiguous to differentiate the candidates by stationarity ratios, we compare the p-values obtained from the ADF test for each sequence. Since a lower p-value indicates a stronger indication of stationarity, we select the parameter combination with a greater number of sequences having lower p-values as the optimal choice.

Additionally, we determine whether to extract seasonality during the decomposition process based on the seasonality ratios. The Exchange, ILI, and Weather datasets have lower seasonality ratios compared to other datasets, indicating that these datasets have a lower significance of seasonality. We decompose these datasets without extracting seasonality.

% Additionally, we examined the significance of the seasonality obtained from the decomposition results. To do so, we performed decomposition using the selected option and compared the scale of seasonality with the trend. As shown in Table ~\ref{tbl:data_info} the seasonality ratio for those datasets are low compared to other datasets. Also, the seasonality scales in the Exchange, ILI, and Weather datasets are smaller than 1/100 of the scale of the trend. 

% This process was repeated for all datasets. The final selected parameters for each dataset were summarized in Table~\ref{tbl:k_p_table}.

% \subsubsection{Normalization} \label{Analysis Normalization}
Fig.~\ref{fig:dist_disc} shows a clear distribution discrepancy between trend components of the training and testing datasets. It was observed in the ETTh1, ETTh2, ETTm2, and ILI datasets. For these datasets, we adopt instance normalization to mitigate distribution discrepancies.
% In such cases, adopting instance normalization might produce good outcomes as it helps overcome the distribution discrepancy problem. Therefore, we use the instance normalization option to adjust the trend for those particular datasets. 
There are less significant changes in the distributions of residual components. Nevertheless, we have observed different variances between the training and testing datasets, and to mitigate these, we apply instance normalization to residual components.
% Even if there are less significant changes in residual distributions, we find out that the training and testing distributions of residual component have different variances. Consequently, it would be successful to make instance normalization possible for such datasets. This enables predictions of residuals in the same distribution, allowing performance improvements. 
For the seasonality component, instance normalization is not applied since seasonality repeats values periodically, as defined.
% On the other hand, we did not apply instance normalization to the seasonality component since it repeats across the whole dataset, according to its definition.}

% \subsubsection{Neural ODEs} 
% \RED{We investigate the benefits of time derivative modeling using the NODE framework. From the perspective of regularization, forecasting with the application of a kinetic regularizer (with kinetic energy regularization coefficient and Jacobian norm coefficient set to 0.1) outperforms the cases without regularization on the ILI and Exchange datasets, making simpler dynamics modeling with regularization more suitable. 
% We also investigate the ODE solver. For combinations of datasets and forecasting horizons that require sophisticated networks, the RK4 method demonstrates superior performance. On the other hand, the Euler method exhibits superior performance in simpler cases.}

\subsection{Ablation Studies}

In order to understand the effectiveness of each block of LTSF-DNODE, we conduct ablation studies on the datasets. 
% According to the analysis in Section~\ref{sec:data_analyses}, there is a distribution discrepancy problem in the ILI, ETTh1, ETTh2, and ETTm2 datasets. Therefore, we perform ablation studies about preprocessing methods on the five datasets.

\begin{table}[t]
    \setlength{\tabcolsep}{3.5pt}
    \caption{Accuracy comparison between LTSF-DNODE and its variants. (Numbers) under datasets are forecasting horizons.}
    % . ``$w/o$. DCMP'' uses the NORM block and the NODE block, ``$w/o$. NORM'' uses the Decomp block and the NODE block, ``$w/o$. NODE'' uses the Decomp block, Norm block, and a single linear layer. LTSF-DNODE uses all blocks. (Numbers) under datasets are forecasting horizons.}
    \label{tbl:abl_table}
    \begin{center}
    \renewcommand{\arraystretch}{0.7}
    \begin{tabular}{ccccccc}    \toprule
        \multirow{2}{*}{\textbf{Models}} &
          \multirow{2}{*}{\textbf{Metrics}} &
          \multicolumn{1}{c}{\multirow{2}{*}{\begin{tabular}[c]{@{}c@{}}\textbf{Exchange}\\ (96)\end{tabular}}} &
          \multicolumn{1}{c}{\multirow{2}{*}{\begin{tabular}[c]{@{}c@{}}\textbf{ILI}\\ (24)\end{tabular}}} &
          \multicolumn{1}{c}{\multirow{2}{*}{\begin{tabular}[c]{@{}c@{}}\textbf{ETTh1}\\ (336)\end{tabular}}} &
          \multicolumn{1}{c}{\multirow{2}{*}{\begin{tabular}[c]{@{}c@{}}\textbf{ETTh2}\\ (336)\end{tabular}}} &
          \multicolumn{1}{c}{\multirow{2}{*}{\begin{tabular}[c]{@{}c@{}}\textbf{ETTm2}\\ (720)\end{tabular}}} \\
         &
           &
          \multicolumn{1}{c}{} &
          \multicolumn{1}{c}{} &
          \multicolumn{1}{c}{} &
          \multicolumn{1}{c}{} &
          \multicolumn{1}{c}{} \\ \midrule
        \multirow{2}{*}{\begin{tabular}[c]{@{}c@{}}$w/o$. DCMP\end{tabular}} &
          MSE &
          0.088 &
          1.662 &
          0.429 &
          0.342 &
          0.366 \\
         &
          MAE &			
          0.206 &
          \textbf{0.844} &
          0.427 &
          0.392 &
          0.382 \\ \midrule
        \multirow{2}{*}{\begin{tabular}[c]{@{}c@{}}$w/o$. NORM\end{tabular}} &
          MSE &
          0.082 &
          1.708 &
          0.433 &
          0.399 &
          0.401 \\
         &
          MAE &
          0.205 &
          0.894 &
          0.434 &
          0.437 &
          0.423 \\ \midrule
        \multirow{2}{*}{\begin{tabular}[c]{@{}c@{}}$w/o$. NODE\end{tabular}} &
          MSE &
          0.082 &
          \textbf{1.620} &
          0.426 &
          0.360 &
          0.365 \\
         &
          MAE &
          0.205 &
          0.893 &
          0.424 &
          0.400 &
          0.383 \\ \midrule
        \multirow{2}{*}{\begin{tabular}[c]{@{}c@{}}LTSF-DNODE\end{tabular}} &
          MSE &
          \textbf{0.078} &
          1.626 &
          \textbf{0.423} &
          \textbf{0.341} &
          \textbf{0.359} \\
         &
          MAE &
          \textbf{0.200} &
          0.848 &
          \textbf{0.422} &
          \textbf{0.390} &
          \textbf{0.381} \\ \bottomrule
        \end{tabular}
    \end{center}
\end{table}

\subsubsection{Decomposition}

% We conducted experiments to verify the effect of the decomposition method (denoted as the DCMP block). We compare the performance of the LTSF-DNODE and ``$w/o.$ DCMP'', which removes decomposition. This allows us to examine the effectiveness of decomposition by comparing the results of these two models. 
To assess the effectiveness of the DCMP block (utilizing the classical decomposition method), we compare the performance of the LTSF-DNODE model with the ``$w/o.$ DCMP'', where the DCMP block is omitted. LTSF-DNODE performs better compared to ``$w/o.$ DCMP'', with a maximum improvement of 11.26\% on MSE and 3.01\% on MAE in the Exchange dataset. As a result, we can infer that time series decomposition has a positive impact on forecasting.

% In addition, we conduct an ablation study for frequency-based transformation methods:
Additionally, we conduct an ablation study to investigate alternative decomposition methods, specifically frequency-based approaches, as follows:
\begin{itemize}
    \item ``FTLinear'' employs Fourier transform~\cite{roberts1987digital} to decompose the time series. It uses low or high-pass filters with frequency criteria of \num{1.00E-04}. The filtered time series passes through individual single linear layers.
    \item ``WTLinear'' employs Wavelet transform~\cite{roberts1987digital} to decompose the time series. This model has a structure similar to ``FTLinear'' and uses a low or high-pass filter.
    % Similar to ``FTLinear'', it uses a low or high-pass filter.
    \item ``TSRLinear'' decomposes time series into trend, seasonality, and residual components using classical methods. This model is identical to the LTSF-DNODE, excluding the NORM and NODE blocks.
\end{itemize}

Table~\ref{tab:various_dcmp} shows the forecasting results of these models. In various benchmark datasets, the classical decomposition method is the better option than other methods.

\begin{table}[t]
    \setlength{\tabcolsep}{4.5pt}
    \caption{MSE and MAE on various decomposition methods}
    \label{tab:various_dcmp}
    \begin{center}
    \renewcommand{\arraystretch}{0.7}
    \begin{tabular}{ccccccc} \toprule
    \textbf{Models}                                                                                  & \textbf{Metrics} & \begin{tabular}[c]{@{}c@{}}\textbf{Exchange}\\ (96)\end{tabular} & \begin{tabular}[c]{@{}c@{}}\textbf{ILI}\\ (24)\end{tabular} & \begin{tabular}[c]{@{}c@{}}\textbf{ETTh1}\\ (336)\end{tabular} & \begin{tabular}[c]{@{}c@{}}\textbf{ETTh2}\\ (336)\end{tabular} & \begin{tabular}[c]{@{}c@{}}\textbf{ETTm2}\\ (720)\end{tabular} \\\midrule
    \multirow{2}{*}{FTLinear}                                                               & MSE     & 0.087                                                   & 1.978                                              & 0.442                                                 & 0.410                                                 & 0.409                                                 \\
                                                                                            & MAE     & 0.215                                                   & 0.983                                              & 0.443                                                 & 0.436                                                 & 0.426                                                 \\\midrule
    \multirow{2}{*}{WTLinear}                                                               & MSE     & 0.086                                                   & \textbf{1.948}                                     & 0.438                                                 & \textbf{0.408}                                        & 0.414                                                 \\
                                                                                            & MAE     & 0.213                                                   & \textbf{0.969}                                     & 0.439                                                 & \textbf{0.435}                                        & 0.429                                                 \\\midrule
    \multirow{2}{*}{TSRLinear}   & MSE     & \textbf{0.079}                                          & 1.981                                              & \textbf{0.437}                                        & 0.422                                                 & \textbf{0.385}                                        \\
                                                                                            & MAE     & \textbf{0.201}                                          & 0.979                                              & \textbf{0.438}                                        & 0.444                                                 & \textbf{0.409}               \\\bottomrule                         
    \end{tabular}
    \end{center}
\end{table}

\subsubsection{Normalization}

To assess how NORM and DENORM blocks impact a dataset with distribution disparities, we compare the performance of LTSF-DNODE with ``$w/o.$ NORM'', which excludes instance normalization. LTSF-DNODE improves performance by up to 14.54\% in MSE and 10.76\% in MAE compared to ``$w/o.$ NORM'' in the ETTh2 dataset.

Therefore, it can be inferred that instance normalization is effective when there exists a distribution discrepancy between the training and testing datasets.

% \begin{figure}[t]
%     \begin{center}
%         \includegraphics[width=0.6\columnwidth]{figures/ode_step.pdf}
%     \end{center}
%     \caption{Exploring the impact of various ODE solver settings. It was conducted for the ILI dataset with the forecasting horizon of 60. DOPRI is represented by a horizontal dashed line since it uses an adaptive step size.}
%     \label{fig:ode_step}
% \end{figure}

\subsubsection{Neural ODEs}

Finally, we investigate the effectiveness of modeling the relationship between the past and the future in LTSF using linear ODEs, as shown in Eq.~\eqref{eq:ODE_base}. We compare two variants: ``$w/o.$ NODE'' and LTSF-DNODE. 
To investigate the effectiveness of NODE, we compare ``$w/o.$ NODE'' and LTSF-DNODE. Both models utilize decomposition and normalization methods. However ``$w/o.$ NODE'' has a single linear layer, while LTSF-DNODE adopts the NODE framework. The experimental results show that LTSF-DNODE outperforms ``$w/o.$ NODE'' with a maximum performance improvement of 5.28\% on MSE and 2.50\% on MAE in the ETTh2 dataset.

\begin{figure}[t]
    \centering
     \begin{center}
        \subfigure[ILI]{\includegraphics[width=0.49\columnwidth]{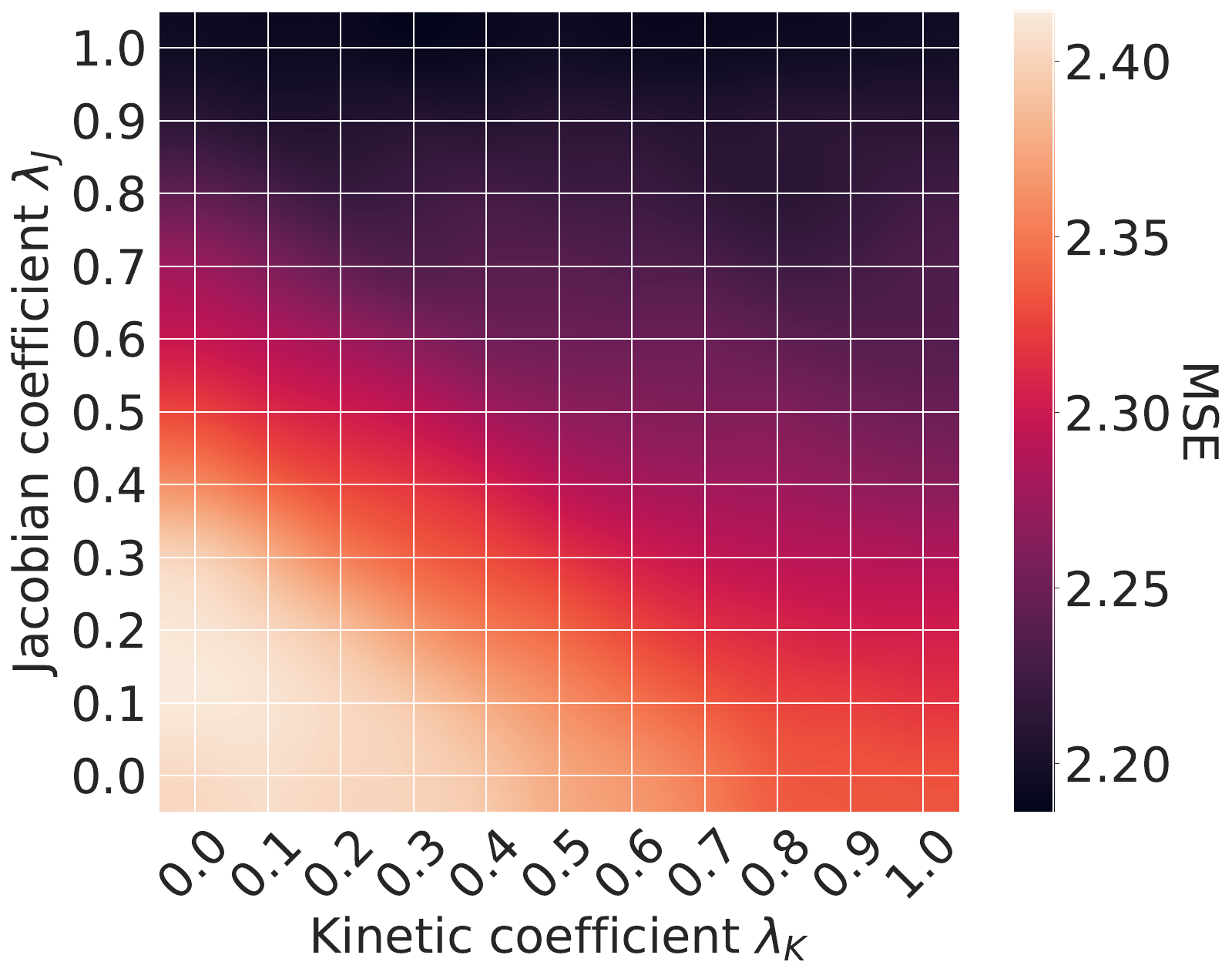}}
        \subfigure[ETTh2]{\includegraphics[width=0.49\columnwidth]{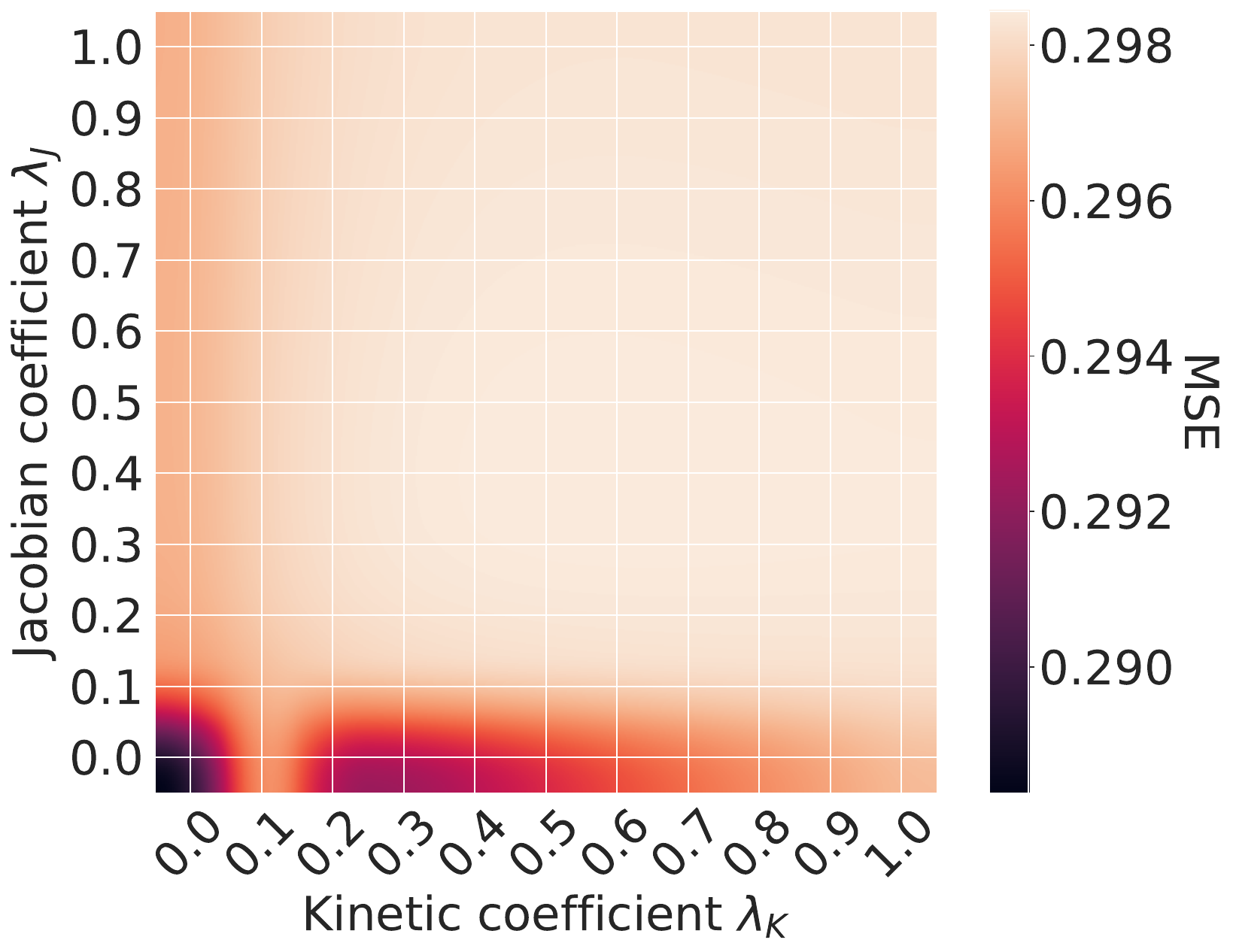}}
    \end{center}    
    \caption{Analyses about the impact of the Jacobian and kinetic regularization on performance (MSE). The forecasting horizon is 60 in (a) and 96 in (b).}
    % We set ODE solver to RK4.}
    \label{fig:regularizer}
\end{figure}

Also, we explore the effect of various learning techniques, such as ODE solver and regularizer, within the NODE framework. The ODE solver is one of the crucial factors in the NODE framework. Table~\ref{tab:ode_step} shows the MSE and MAE against various ODE solvers.
%The loss decreases as the step size gets smaller, which shows that a smaller step size can produce more accurate forecasting.
Additionally, the Jacobian and kinetic regularizers enable smoother learning. It transforms unstable learning into stable learning, leading to an improvement in forecasting accuracy. In Fig.~\ref{fig:regularizer} (a), we can see the loss decreases as the Jacobian and kinetic coefficients increase, respectively; however, in Fig.~\ref{fig:regularizer} (b), it does not. These findings indicate that the effectiveness of the regularizer depends on the dataset. As a result, the incorporation of a regularizer suggests its potential to enhance the learning process of our proposed model.

\begin{table}[t]
    \caption{MSE and MAE on various ODE solvers. It was conducted for the ILI dataset with the forecasting horizon of 60.}
    \label{tab:ode_step}
    \begin{center}
        \begin{tabular}{ccc} \toprule
        \textbf{ODE Solver} & \textbf{MSE}   & \textbf{MAE}   \\\midrule
        Euler      & 1.836 & 0.942 \\
        RK4        & 1.693 & 0.911 \\\bottomrule 
        % Dopri      &       &        \\\bottomrule 
        \end{tabular}
    \end{center}
\end{table}

\subsubsection{Model Efficiency Analyses}
\begin{figure}[t]
    \begin{center}
        \includegraphics[width=0.6\columnwidth]{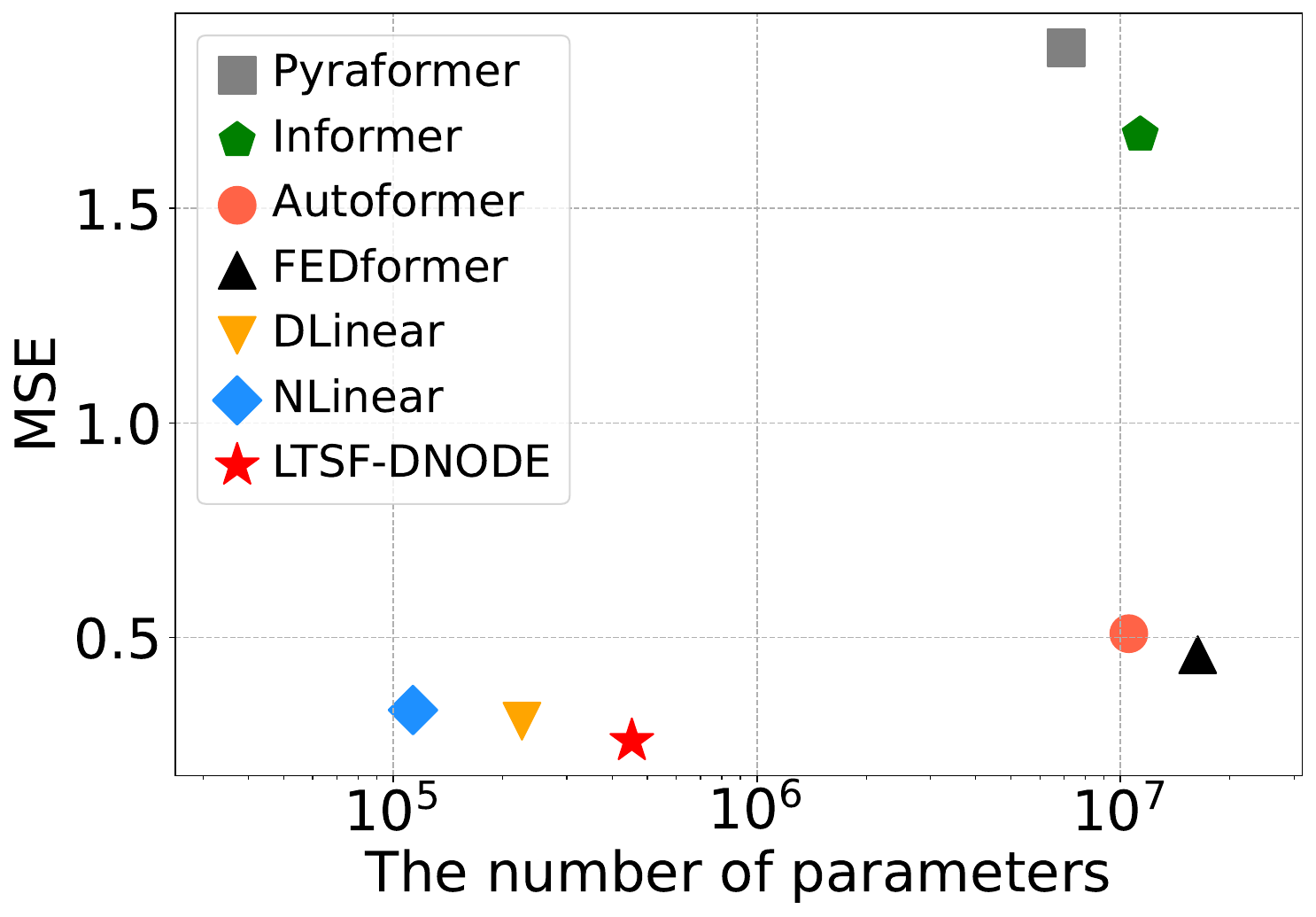}
    \end{center}
    \caption{Model efficiency. It was measured for the Exchange dataset. The bottom left corner is preferred.}
    \label{fig:param}
\end{figure}

Fig.~\ref{fig:param} shows the number of parameters and the MSE of models. Our LTSF-DNODE model employs fewer parameters compared to Transformer-based models and slightly more parameters compared to Linear-based models. However, LTSF-DNODE outperforms these models on various datasets.

\section{Related Work}

Recently, various Transformer-based models have shown encouraging results in the field of LTSF. Transformer-based LTSF models overcome the limitations of RNNs in LTSF by enabling direct multi-step forecasting, but difficulties remain due to quadratic time complexity, high memory usage, and the inherent limitations of the encoder-decoder architecture.

Transformer-based LTSF models have addressed these issues in various ways. LogTrans~\cite{li2019enhancing} utilizes a convolutional self-attention mechanism to address challenges related to locality-agnostic property and memory limitations. Pyraformer~\cite{liu2021pyraformer} uses an interscale tree structure, reflecting a multi-resolution representation of a time series dataset. Informer~\cite{zhou2021informer} introduces a Probsparse self-attention to minimize time complexity and reduce the processing time. Also, it uses a generative decoder to improve performance.

On the other hand, other transformer-based models have attempted to learn the characteristics of time series by combining self-attention structures and decomposition methods. Autoformer~\cite{wu2021autoformer} aims to enhance forecasting accuracy by capturing auto-correlation and employing the decomposition method, which extracts trend and seasonality. This structure enables progressive decomposition capabilities for intricate time series. FEDformer~\cite{zhou2022fedformer} also utilizes the Fast Fourier transform with a low-rank approximation for preprocessing the temporal data. Moreover, it employs a mixture of expert decomposition, which extracts trend and seasonality, to manage the distribution discrepancy problem.

Due to the issues with Transformers, non-Transformer-based models have also emerged. N-BEATS~\cite{oreshkin2019n} incorporates fully-connected networks with trend and seasonality decomposition to enhance interpretability. Furthermore, DEPTS~\cite{fan2022depts} advances N-BEATS to propose a more effective model that specializes in learning periodic time series. SNaive~\cite{li2022simpler} uses simple linear regression and demonstrates that a simple statistical model using linear regression could achieve better performance.  NLinear and DLinear~\cite{zeng2023transformers} are Linear-based models that use a single linear layer with simple preprocessing. NLinear utilizes a normalization method that subtracts the last value of the sequence from the input. DLinear utilizes the classical decomposition method that decomposes the input into a trend and a remainder.

\section{Conclusion and Future Work}

Long-term time series forecasting is an important research topic in deep learning, and simple models such as Linear-based approaches are showing good performance. However, these models are too simple to represent the dynamics of the time series. Our proposed method, LTSF-DNODE, uses a neural ODE (NODE) framework with a simple architecture and utilizes decomposition depending on data characteristics. Our contribution allows the model to not only use temporal information appropriately but also capture time series dynamically. We experimentally demonstrated that LTSF-DNODE outperforms the existing baselines on real-world benchmark datasets. In ablation studies, we analyzed how the main components of LTSF-DNODE affect performance.

In future work, we will refine the modeling of the residual component. Our analysis results demonstrate the evident stationarity of the residual component within the optimal settings we have identified. Therefore, we hypothesize that the performance of LTSF can be further improved through an elaborate modeling of the residual component, potentially via an approach such as stochastic differential equations (SDEs).

\section*{Acknowledgement}
Noseong Park is the corresponding author. This work was supported by the Institute of Information \& Communications Technology Planning \& Evaluation (IITP) grant funded by the Korean government (MSIT) (No. 2020-0-01361, Artificial Intelligence Graduate School Program at Yonsei University, 10\%), and (No.2022-0-00857, Development of AI/data-based financial/economic digital twin platform, 90\%)

\bibliographystyle{IEEEtran}
\bibliography{LTSF-DNODE}

\end{document}